\documentclass[conference]{IEEEtran}
\IEEEoverridecommandlockouts
\usepackage{cite}
\usepackage{amsmath,amssymb,amsfonts}
\usepackage{algorithmic}
\usepackage{graphicx}
\usepackage{textcomp}
\usepackage{xcolor}
\usepackage{url}
\usepackage{booktabs}
\usepackage{pgfplots}
\pgfplotsset{compat=1.18}
\usepackage[caption=false]{subfig}
\usepackage{multirow}
\usepackage{booktabs}
\usepackage{tabularx}
\bstctlcite{IEEEexample:BSTcontrol}

\def\BibTeX{{\rm B\kern-.05em{\sc i\kern-.025em b}\kern-.08em
    T\kern-.1667em\lower.7ex\hbox{E}\kern-.125emX}}

\begin{document}

% --- TITLE ---
\title{Cutting the Cord: System Architecture for Low-Cost, GPU-Accelerated Bimanual Mobile Manipulation\\
\thanks{We thank the original developers for creating Xlerobot, which enabled this study.}
}

%  TEMPORARILY REDACTED FOR DOUBLE BLIND. DO NOT REMOVE !!!

\author{\IEEEauthorblockN{Artemis Shaw, Chen Liu, Justin Costa, Rane Gray, Alina Skowronek, Kevin Diaz, Nam Bui, Nikolaus Correll}
}

 \maketitle

% --- ABSTRACT: ---
\begin{abstract}
We present a bimanual mobile manipulator built on the open-source XLeRobot with integrated onboard compute for less than \$1300. Key contributions include: (1) optimized mechanical design maximizing stiffness-to-weight ratio, (2) a Tri-Bus power topology isolating compute from motor-induced voltage transients, and (3) embedded autonomy using NVIDIA Jetson Orin Nano for untethered operation. The platform enables teleoperation, autonomous SLAM navigation, and vision-based manipulation without external dependencies, providing a low-cost alternative for research and education in robotics and robot learning. Design files and documentation are available at \url{double blind}.
\end{abstract}


\section{Introduction}
There exists a wide range of low-cost, open-source platforms for robotics research and education \cite{vrochidou2018open,curriculum, vega2025g,bindu2019cost, cone-e}. Recent advances in vision-language-action (VLA) models and imitation learning have renewed interest in affordable hardware platforms capable of supporting data collection and policy deployment. Low-cost manipulators such as the ViperX/WidowX arms have been used in bimanual teleoperation setups \cite{act_aloha, aldaco2024aloha}, including mobile variants \cite{mobile_aloha} at significantly higher system cost. At the same time, increased interest in humanoid robotics is driving interest in an affordable and safe dual-arm mobile manipulation platform that can serve as a low-barrier baseline to compare capabilities and cost. 

The Hugging Face ``LeRobot'' ecosystem \cite{lerobot, lekiwi} has lowered the entry barrier by providing open-source designs for both manipulation and mobility, recently combined into the XLeRobot platform \cite{xlerobot}. In its reference configuration, XLeRobot offers a tethered dual-arm mobile manipulator for less than \$700 in parts, making it one of the most accessible bimanual mobile systems currently available.

%\begin{figure}[!t]
%\centering
%\includegraphics[width=\columnwidth]{pictures/xlerobot.jpg}
%\caption{The evolved XLeRobot platform performing a %manipulation task.}
%\end{figure}

Yet, practical untethered deployment of low-cost mobile manipulators remains challenging. In particular, building systems based on online instructions introduces non-trivial systems engineering problems that are often under-documented. These include structural compliance in 3D-printed links, voltage instability under high transient motor loads, power distribution constraints, and thermal limits of embedded compute. While individually well understood, their interaction in tightly integrated, resource-constrained mobile manipulators can lead to difficult-to-debug, hard-to-explain problems such as unpredictable resets of servos, control boards, or on-board compute, reduced payload capacity, and premature actuator wear.
\begin{figure}[!t]
\centering
\includegraphics[width=0.9\columnwidth]{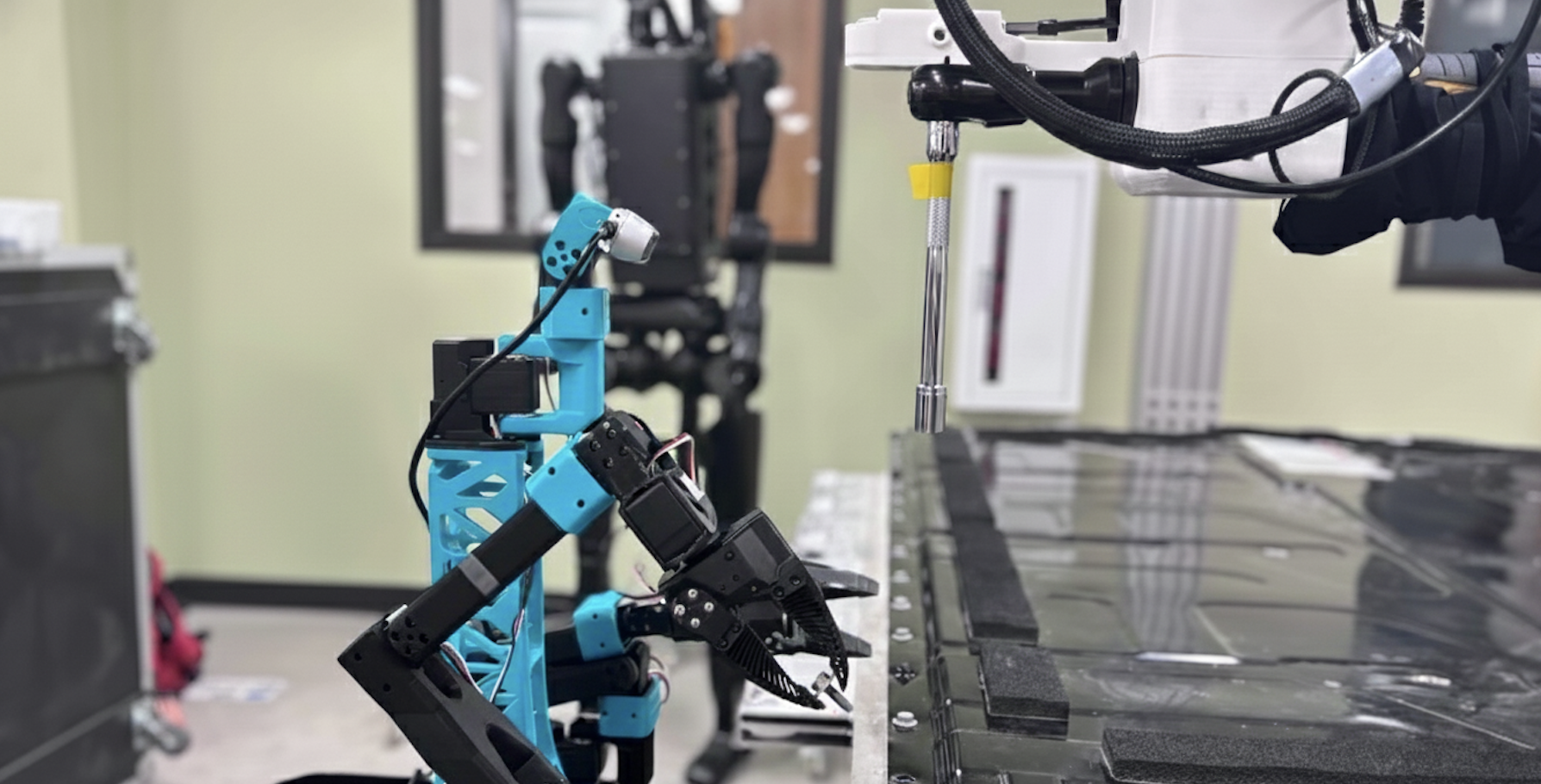}
\caption{The evolved XLeRobot platform a low-cost bimanual platform for education and research assisting an industrial robot by removing screws from an electric vehicle battery. }
\label{fig:battery_disassembly}
\end{figure}
For example, insufficient conductor sizing and daisy-chained power rails can produce voltage collapse during concurrent multi-axis actuation and specific load patterns. Similarly, compliant 3D-printed links can act as unintended series-elastic elements \cite{series_elastic_actuator}, reducing effective torque transmission and increasing drivetrain stress. Off-the-shelf Lithium-Polymer battery packs provide high energy density, but require careful handling and protection circuitry. Commercial power stations offer a safer alternative, but impose strict current limits that must be respected at both the hardware and firmware levels. Finally, faulty start-up scripts might lead to brown-outs for some initial poses when all servos start up at once.

This paper addresses these practical but critical integration challenges through a rigorously characterized, untethered evolution of the XLeRobot platform (Fig. \ref{fig:battery_disassembly}). % We integrate onboard NVIDIA Jetson Orin Nano Super compute \cite{jetson_orin_nano_super} within a system-level redesign focused on structural efficiency, power stability, and reproducible autonomy. The resulting platform operates fully untethered at a total cost of approximately \$1300 and supports onboard inverse kinematics, RGB-D perception, SLAM-based navigation, and teleoperation. The system is designed to support modern learning-based policies and GPU-accelerated inference, and we benchmark its current computational limits under representative workloads.

\begin{figure*}[t]
    \centering
    \includegraphics[width=\textwidth]{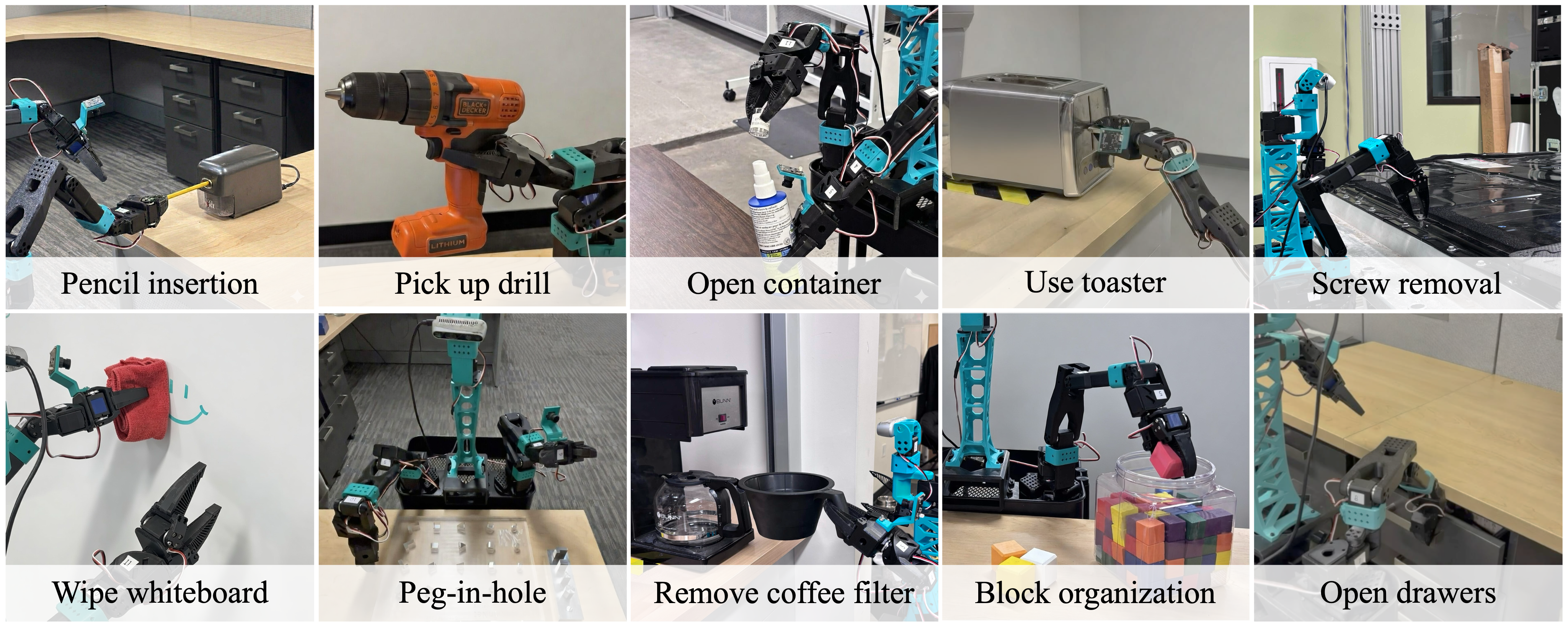}
    \caption{Diverse Bimanual Manipulation Tasks. Demonstrations include pencil insertion into an electric pencil sharpener, picking up a YCB drill \cite{calli2017yale}, bimanual container opening, toaster use, screw removal from car-battery disassembly, whiteboard wiping, peg-in-hole insertion, coffee filter removal, block organization, and drawer opening.}
    \label{fig:task_demos}
\end{figure*}

Our contributions are primarily at the systems level, resulting in an entry-level platform for research and education:

\begin{itemize}
    \item \emph{Optimized Structural Topology:} A graded 3D-printing strategy that maximizes stiffness-to-weight ratio, increasing functional load capacity while reducing mass.
    
    \item \emph{Tri-Bus Power Topology:} A power distribution architecture that electrically isolates high-transient motor loads from embedded compute, combined with firmware-level current envelope to prevent brownouts under untethered operation.
    
    \item \emph{Embedded Autonomy Stack:} A reproducible ROS2-based architecture integrating onboard perception, inverse kinematics, SLAM, navigation, and teleoperation, with benchmarking of edge compute performance using a Jetson Orin.
    
    \item \emph{Open and Reproducible Design:} Complete CAD, electrical schematics, characterization data, and documentation to facilitate replication in research and education.
\end{itemize}

% --- SECTION II: RELATED WORK ---
\section{Related Work}
%In recent years, low-cost mobile manipulation solutions have gained significant interest for education and research. They enable scalable data collection for embodied AI and reduce the financial barrier to entry for hands-on robotics training. Here, the focus has shifted from single-arm mobile manipulators \cite{correll2010peer,amsters2019turtlebot} toward dual-arm mobile manipulators that serve as accessible stand-ins for full-scale humanoid systems and commercial-grade platforms for research in imitation learning and bimanual coordination. 
Table \ref{tab:comparison} summarizes existing dual-arm mobile manipulators and their cost. 

\begin{table}[!htb]
\centering
\caption{Comparison of Dual-Arm Mobile Manipulators}
\label{tab:comparison}
\small
\renewcommand{\arraystretch}{1.25}
\resizebox{\columnwidth}{!}{%
\begin{tabular}{@{}l c c c c c c@{}}
\toprule
\textbf{Platform} & \textbf{Compute} & \textbf{GPU} & \textbf{Untethered} & \textbf{DoF} & \textbf{Payload} & \textbf{Est.\ Cost} \\
\midrule
AhaRobot \cite{aharobot}
  & Mini-ITX (i5)
  & RTX\,4060
  & Yes
  & 16
  & $\approx$\,1.5\,kg/arm
  & $\approx$\,\$2k \\
Cone-E \cite{cone-e, agilex_piper}
  & Intel NUC
  & ---
  & Yes
  & 16
  & $\approx$\,1.5\,kg/arm
  & $<$\,\$12k \\
Mobile ALOHA \cite{mobile_aloha}
  & Laptop (onboard)
  & RTX\,3070\,Ti
  & Yes
  & 16
  & $\approx$\,0.75\,kg/arm
  & $\approx$\,\$32k \\
TB3 Waffle Pi + arms \cite{amsters2019turtlebot, robotis_open_x}
  & Raspberry Pi 4
  & ---
  & Yes
  & 12
  & $\approx$\,0.5\,kg/arm
  & $\approx$\,\$3k \\
XLeRobot \cite{xlerobot}
  & External PC
  & ---
  & No
  & 15
  & $\approx$\,0.8\,kg
  & $\approx$\,\$700--1k \\
\textbf{Ours}
  & \textbf{Jetson Orin}
  & \textbf{Ampere}
  & \textbf{Yes}
  & \textbf{17}
  & \textbf{1.0\,kg/arm}
  & $\boldsymbol{\approx}$\,\textbf{\$1.2k} \\
\bottomrule
\end{tabular}%
}

\vspace{0.4em}
\raggedright
\footnotesize
\footnotesize{ \small \emph{Note:} DoF includes all independently actuated joints.
Costs and specifications based on 2025--2026 public BOMs and publications, although we could not corroborate the cost of the AhaRobot.}

\end{table}

While the xLeRobot has inspired this project, achieving true autonomy requires integrating sensing and computation into the chassis. The AhaRobot \cite{aharobot} utilize Raspberry Pi units for onboard compute, but these setups face a significant throughput ceiling when executing high-frequency inference for Vision-Language Models (VLMs), as they lack a Graphics Processing Unit (GPU). The use of CUDA-accelerated edge modules, such as the NVIDIA Jetson Orin Nano \cite{jetson_orin_nano_super}, has emerged as a middle-ground solution. However, this approach introduces significant engineering trade-offs regarding energy density and power-bus stability \cite{energy_sources}. 

The evolved xLeRobot aims at providing minimal requirements for untethered operation that is capable of research and education in robotics and robotic learning, i.e. providing the power infrastructure to support a GPU-enabled on-board computer, as well as providing the workspace and height that allows to perform meaningful tasks.

% --- Materials and Methods/Hardware architecture ---
\section{Materials and Methods}
The xLeRobot combines the LeKiwi omni-directional drive kinematics \cite{lekiwi} with two 3 degrees-of-freedom (DoF), dual SO-101 manipulators \cite{so-101} with 5+1 DoF (one DoF for the gripper), a 2-DoF neck, and a hybrid structural architecture integrating a mass-manufactured steel utility substrate (IKEA R{\AA}SKOG) with custom FDM interfaces, reaching a height of 1.20m. %This offers steel rigidity for load-bearing and consumer-grade price, while using 3D printing for complex geometric connections and component housing. 
To support sustained untethered autonomy, the NVIDIA Jetson Orin Nano is embedded in the chassis mid-plane using a custom enclosure with passive thermal ducting. An Intel RealSense D435 on a 2-DOF neck provides RGB-D data for SLAM navigation and manipulation policies. Table \ref{tab:bom} details the system components.

\begin{table}[h!]
\caption{\emph{System Bill of Materials (BOM)}}
\vspace{-15px}
\label{tab:bom}
\begin{center}
\resizebox{\columnwidth}{!}{%
\begin{tabular}{|l|l|c|r|}
\hline
\textbf{Subsystem} & \textbf{Component} & \textbf{Qty} & \textbf{Cost (USD)} \\
\hline
\textbf{Compute} & NVIDIA Jetson Orin Nano Super \cite{jetson_orin_nano_super}& 1 & \$249.00 \\
\hline
\textbf{Compute} & microSD Card & 1 & \$11.23 \\
\hline
\textbf{Peripherals} & Anker 4-in-1 USB-C Hub & 1 & \$14.99 \\
\hline
\textbf{Actuation} & Feetech STS-3215 Servos (12V) \cite{feetech}& 17 & \$271.83\\
\hline
\textbf{Actuation} & Servo Wire: 50' 3-Color & 1 & \$15.97 \\
\hline
\textbf{Actuation} & Wonrabai Serial Bus Servo Driver Board & 2 & \$21.10 \\
\hline
\textbf{Power} & Anker SOLIX C300 Power Station \cite{anker} & 1 & \$159.99 \\
\hline
\textbf{Structure} & PLA Filament (3kg) & 1 & \$45.00 \\
\hline
\textbf{Structure} & M3 Screws and Nuts Set & 1 & \$14.99 \\
\hline
\textbf{Frame} & IKEA RÅSKOG Utility Cart & 1 & \$39.99 \\
\hline
\textbf{Mobility} & 4'' Omni Wheels & 3 & \$29.97 \\
\hline
\textbf{Perception} & Intel RealSense D435\textsuperscript{*} & 1 & \$333.75 \\
\hline
\textbf{Power Cable} & USB-C to DC5521 Cable & 1 & \$8.99 \\
\hline
\textbf{Power Cable} & USB-C to DC5525 PD 140W Cable & 1 & \$11.99 \\
\hline
\textbf{Power Cable} & DC5521 Car Cigarette Lighter Cable & 1 & \$9.49 \\
\hline
\textbf{Data Cable} & USB-C to USB-C Cable (2 pcs) & 1 & \$8.99 \\
\hline
\hline
\textbf{TOTAL} & \textbf{Untethered xLeRobot} & \textbf{-} & \textbf{\$1202.28} \\
\hline
\end{tabular}%
}
\end{center}
%\begin{flushleft}
%\small \textsuperscript * The system is designed to be compatible with various USB depth cameras; pricing may vary depending on the chosen model.
%\end{flushleft}
\end{table}

\subsection{Structural Design and Fabrications}

\begin{figure}[!htb]                
    \centering                
    \includegraphics[width=0.4\columnwidth]{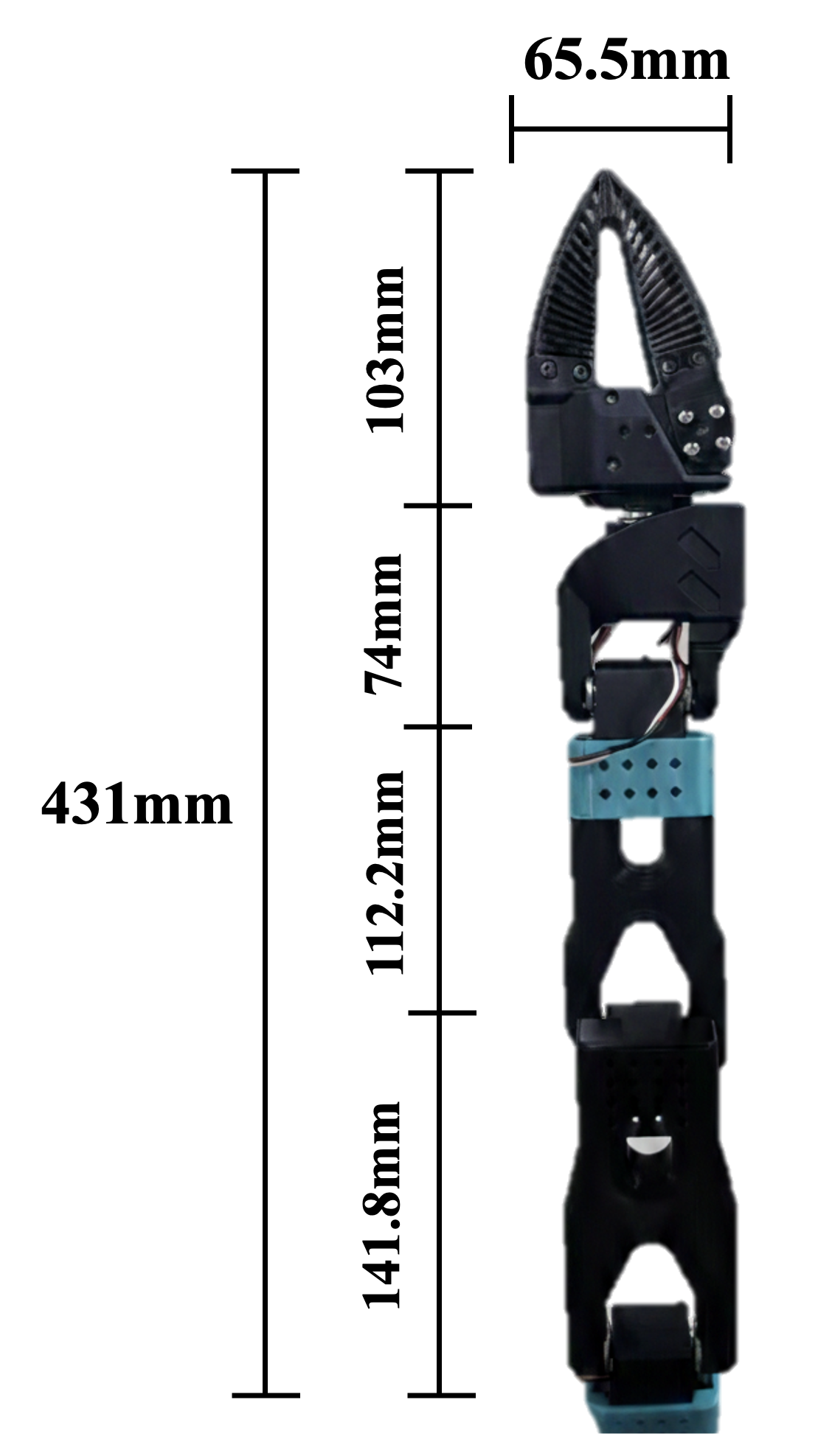}                
    \includegraphics[width=0.5\columnwidth]{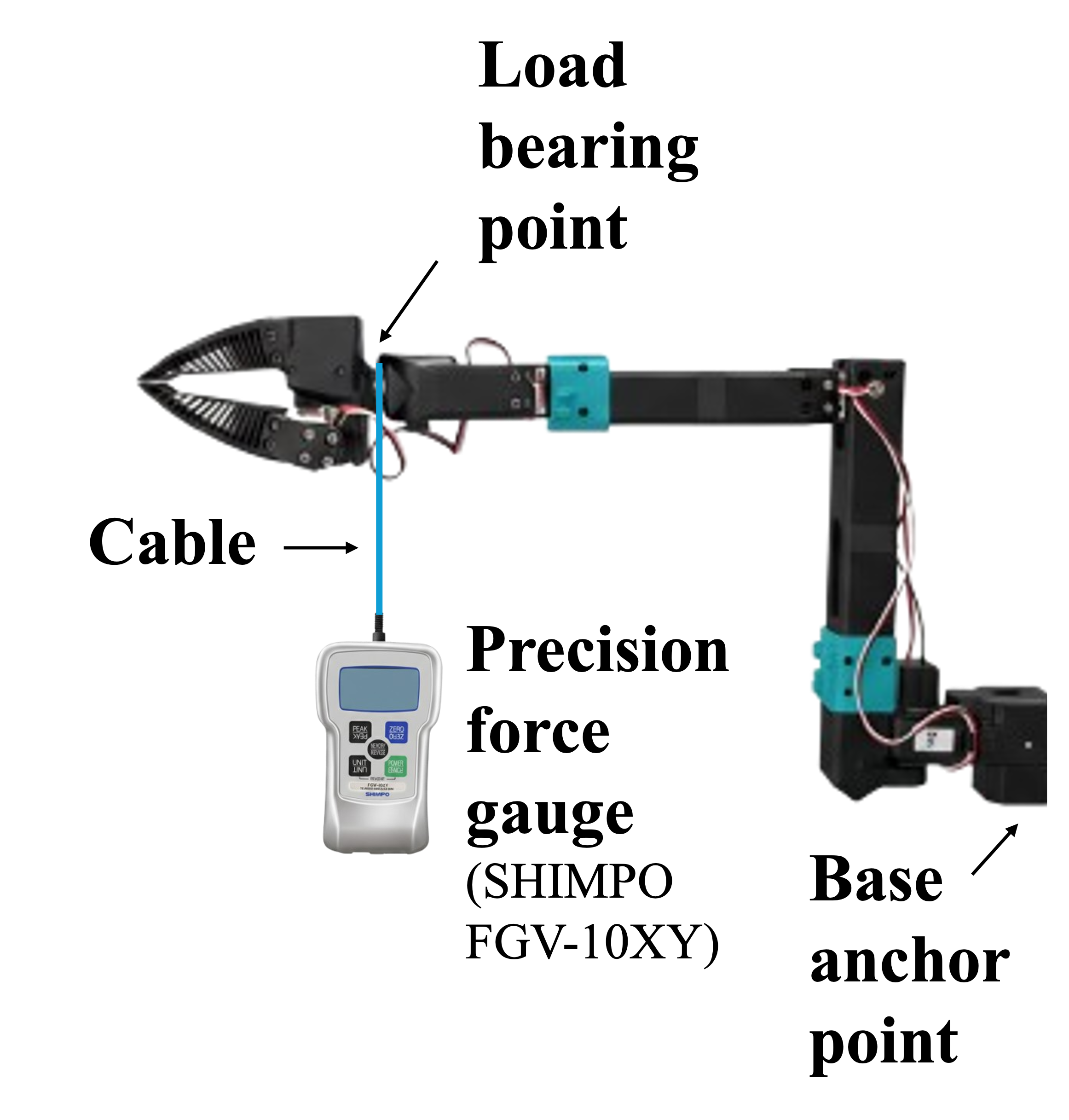}    
    \caption{Structural composition and kinematics: (left) Robotic arm with 4-wall (black)/2-wall topology (cyan), and (right) cantilever experimental setup.}      
    \label{fig:robotic_arm_full}                
\end{figure}

Standard FDM profiles (2 perimeters, 15\% infill, Polyactic Acid, PLA, Ouverture, 1.75mm) exhibit excessive compliance under torsional loads. We employ functionally graded topology: a High-Shell topology (4 perimeters, 15\% infill) for primary links to maximize the Second Moment of Area ($I$) \cite{ mazlan_3d_printing_effects}. This approach provides a rigid seat for actuators, preventing motor misalignment. Conversely, compliant components, such as the joint belts stabilizing the motors, use the Baseline topology (2 perimeters, 15\% infill) for assembly flexibility. Fig. \ref{fig:robotic_arm_full} shows an arm along with the dimensions of individual links, with an overall reach over 40cm. 

A primary goal of our topology is to maximize transmission efficiency by increasing the stiffness $k$ of each link. A compliant link (low $k$) acts as a series elastic actuator \cite{series_elastic_actuator} and absorbs motor work as elastic potential energy, defined by:
\begin{equation}
    U = \frac{1}{2}k\delta^2
\end{equation}
where $\delta$ represents structural deflection. This parasitic compliance wastes torque, requiring the motor to reach its stall threshold ($\tau_{max}$) at a lower external force. By implementing rigid 4-wall links, we minimize $\delta$, converting torque directly into end-effector force. This allows the SO-101 to utilize the full torque profile of its actuators---enabling a 1~kg payload capacity---while protecting internal gear-trains from fatigue.

Additionally, following the original design \cite{xlerobot} the platform utilizes a compliant gripper geometry based on the Fin Ray effect---a bio-inspired principle leveraging passive morphological computation (Fig. \ref{fig:robotic_arm_full}) printed in Thermoplastic Polyurethane (TPU, Ouverture, 1.75mm). The triangular rib-and-strut geometry allows the fingers to wrap around objects when force is applied, relying on controlled material deformation rather than complex sensor feedback.

\subsection{Power Distribution and Tri-Bus Topology}
\begin{figure}[h!]
\centering
\includegraphics[width=\columnwidth]{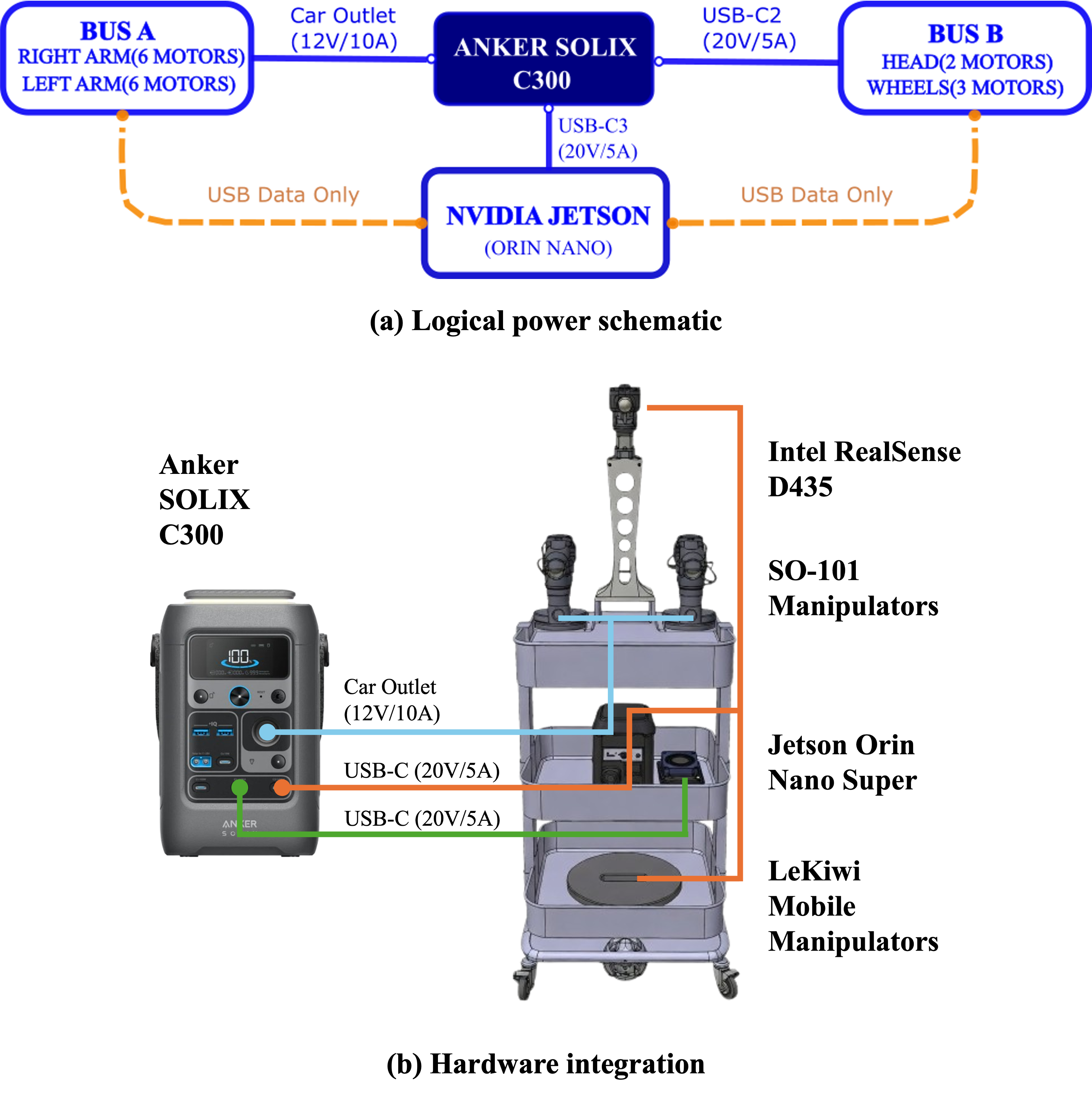}
\caption{Power System Architecture. (a) The logical power schematic illustrates the distribution from the Anker SOLIX C300 to the distinct motor buses and the NVIDIA Jetson controller via high-output USB-C and car outlet interfaces. (b) The hardware integration diagram shows the physical routing of power lines within the mobile manipulator chassis, supporting the Intel RealSense D435 and dual arm configurations. }
\label{fig:power_distribution}
\end{figure}
The system uses an Anker SOLIX C300 (288Wh/300W) \cite{anker} Power Distribution Unit (PDU) to provide robust over-current protection while avoiding custom LiPo assembly risks \cite{xlerobot}. Proper power distribution is crucial for mobile manipulators; simultaneous high-torque movements generate significant voltage transients that can induce compute brownouts or system resets if the Jetson Orin Nano is not properly isolated from the actuator power rails. Quantitative testing on the original daisy-chained reference design showed that the 3A and 5A current ceilings per port were insufficient to handle peak dynamic loads. Qualitatively, this manifested as frequent compute shutdowns during bimanual manipulation.

To decouple high-torque loads from the compute domain, we implemented a Tri-Bus Topology (Fig. \ref{fig:power_distribution}). 
The SOLIX C300 features three USB-C ports (two 140W, one 100W) and a 12V DC car outlet capable of 10A. We consolidate the high-draw dual manipulators onto the 10A DC car outlet. The wheels and neck actuator bus (Bus A) are powered by a high-power USB-C rail, while the Jetson Orin Nano is isolated on a separate high-power USB-C rail. 
The system employs $N=17$ Feetech STS3215 servos with a stall current ($I_{stall}$) of $2.7$\,A \cite{feetech}. We model the load $I(\tau, \alpha)$ as a function of Torque ($\tau$) and Acceleration ($\alpha$) to enforce a software-defined Safe Operating Envelope (SOE):
\begin{equation} \label{eq:current_model}
I_{load}(\tau, \alpha) = N_{a} \cdot \left[ \underbrace{\left(2.52 \cdot \frac{\tau}{1000}\right) + 0.18}_{\text{Static Load Component}} + \underbrace{D(\alpha)}_{\text{Dynamic Inrush}} \right]
\end{equation}

Here, $0.18$A represents the baseline no-load running current, and $2.52$A is the net current slope required to reach the rated stall torque at $2.7A$ \cite{feetech}.  The dynamic inrush $D(\alpha)$ is mitigated using the STS3215’s internal acceleration registers.

To prevent PDU shutdown, we derived firmware-level saturation limits based on the specific constraints of the SOLIX C300. Maximum torque $\tau$ is calculated by isolating it from the current model $I_{load}$ based on the per-port limit $I_{L}$:

\begin{equation}
\tau \leq \left( \frac{\frac{I_{L}}{N_{a}} - 0.18 - D(\alpha)}{2.52} \right) \times 1000
\end{equation}

In these models, $I_{L}$ represents the maximum current limit of the specific PDU port, and $N_{a}$ represents the number of active actuators on that bus. Table \ref{tab:final_settings} summarizes these hard-coded limits, acting as a ``virtual fuse''. With actuators capped at $\approx 240$W ($20$A at $12$V), the system maintains $60$W headroom for the Jetson Orin Nano, which requires between 15W and 25W depending on inference load, ensuring zero-latency computation without risk of brownouts \cite{jetson_orin_nano_super}.

\begin{table}[h!]
\centering
\caption{Firmware Saturation Settings}
\label{tab:final_settings}
\renewcommand{\arraystretch}{1.3}
\resizebox{\columnwidth}{!}{
\begin{tabular}{|l|c|c|c|c|l|}
\hline
\textbf{Bus Group} & \textbf{Torque} & \textbf{Accel} & \textbf{Load} & \textbf{Limit} & \textbf{PDU} \\
 & \textbf{($\tau$)\textsuperscript{*}} & \textbf{($\alpha$)} & \textbf{($I_{total}$)} & \textbf{(Fuse)} & \textbf{Constraint} \\ \hline
\hline
\textbf{Bus A (Wheels/Neck)}  & 650 & 20 (Med) & 9.84 A & 10 A & USB-C2 Output \\ \hline
\textbf{Bus B (Arms)}  & 450 & 40 (Soft)  & 4.90 A & 5 A & DC Car Outlet \\ \hline
\textbf{Jetson Orin}   & N/A & N/A       & $<$2.1 A & 5 A & USB-C3 Output \\ \hline
\end{tabular}
}
\begin{flushleft}
\small \textsuperscript * Values for $\tau$ represent the integer settings within the motor firmware registers ($0$--$1000$). Torque Conversion: $\tau=450 \approx 1.32$\,N$\cdot$m; $\tau=650 \approx 1.91$\,N$\cdot$m.
\end{flushleft}
\end{table}

% --- SECTION V: SOFTWARE & CONTROL ARCHITECTURE ---
\section{SOFTWARE \& CONTROL ARCHITECTURE}
The platform uses a hierarchical control stack with ROS 2, Pinocchio/Pink \cite{carpentier-sii19,pink}, and Open3D.

%\subsection{Embedded Compute Platform}
The NVIDIA Jetson Orin Nano Super provides 67 INT8 TOPS and 102 GB/s memory bandwidth within 7--25W \cite{jetson_orin_nano_super}, sufficient for vision backbones (YOLO) \cite{sam}, quantized LLMs (Llama 3.1) \cite{llama31}, transformer policies (ACT, Diffusion Policy) \cite{act_aloha, diffusion_policy} and TensorRT accelerates inference \cite{tensorrt}.

\begin{figure}[h!]                
    \centering
    \includegraphics[width=0.8\linewidth]{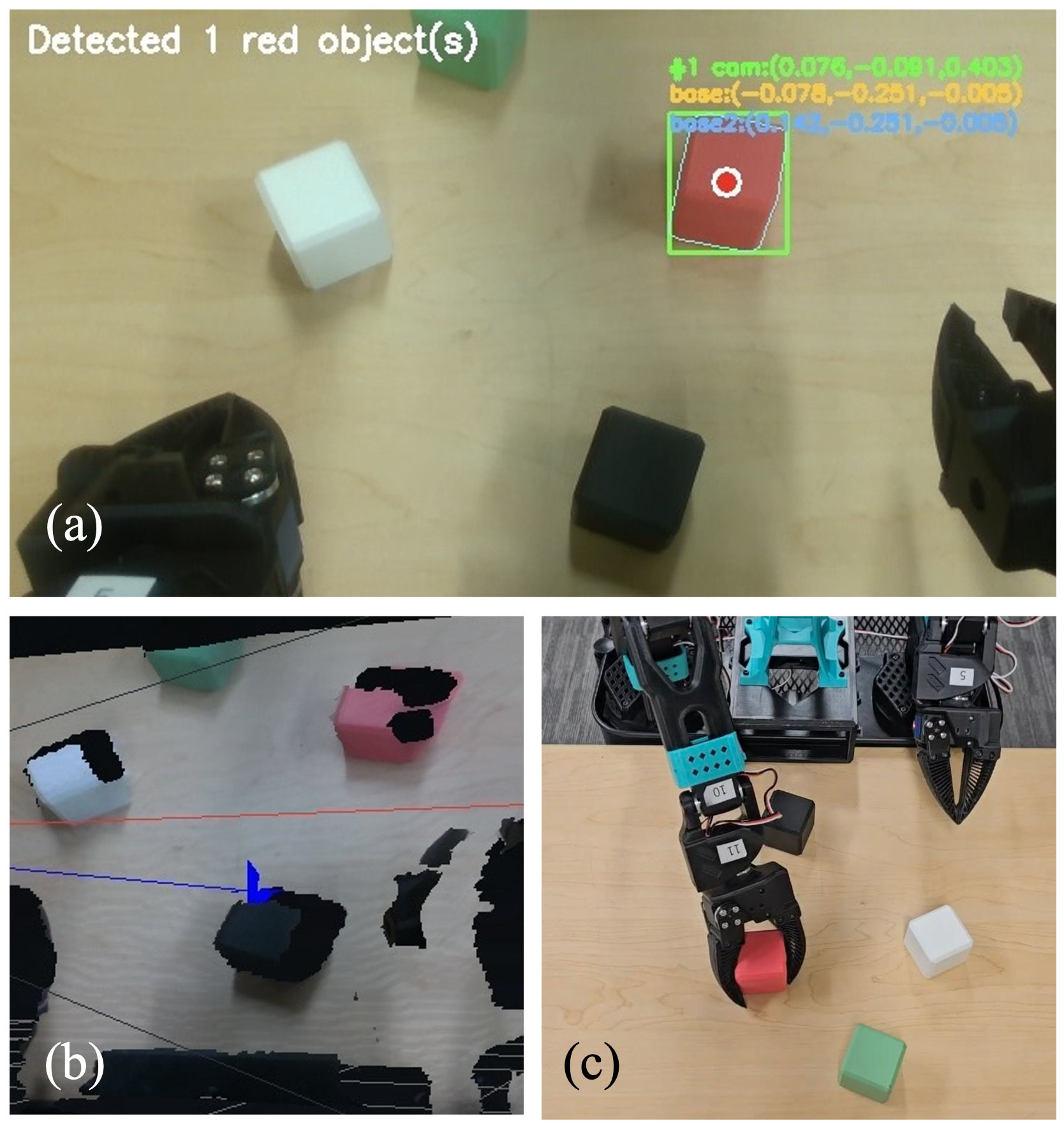}    
    \caption{Perception-driven manipulation pipeline. (a) Color-based segmentation and centroid detection in the camera optical frame. (b) 3D mapping and representation of the environment. (c) Final grasp execution of the target object following inverse kinematics planning.}                
    \label{fig:grasping_pipeline}                
\end{figure}

\subsection{Task-based inverse kinematics and manipulation}
In order to support classical robot control policies, we implemented a deterministic, perception-driven manipulation pipeline that converts color and depth (RGB-D) into synchronized motor commands. The pipeline is structured as a sequence of modular, independently verifiable stages: perception, coordinate transformation, task-based inverse kinematics (IK), and motor mapping (Fig. \ref{fig:grasping_pipeline}).

The process begins by capturing an RGB-D frame from the Intel RealSense D435. An object is identified via a color-based detector---implemented through Hue, Saturation, Value (HSV) thresholding and connected-component analysis---which computes the 3D centroid of the target label (e.g., ``red'') in the camera's optical frame (Fig. \ref{fig:grasping_pipeline}a).

This target point is then transformed into the SO-101 arm base frame using the robot's MuJoco's MJCF kinematic model. The transformation accounts for the current head joint angles to obtain the instantaneous camera pose, applies the required coordinate alignment, and adds a small, hand-tuned offset in the base frame to compensate for systematic calibration errors.

Task-based IK is available via the pink \cite{pink} library. It formulates the task-based IK problem as a quadratic program (QP) with constraints. Given the target pose for the left or right arm and posture constraints for the additional degrees of freedom, pink will generate a joint-space trajectory $\{q_0, \dots, q_T\}$ at $100$\,Hz ($dt = 0.01$\,s). The solver minimizes end-effector position error while applying posture regularization to avoid extreme configurations. By choosing different gains for frame tasks and posture constraints, the system will automatically trade between the mobile base and using the arm kinematics. Similarly, by restricting the movement of the opposite arm during a grab, pink is able to avoid self-collisions with the other arm. Yet, formulating IK as a constrained QP is not guaranteed to find a solution, which will require motion planning. 

The trajectory is streamed as synchronized position commands. Upon reaching the target pose, a stepped gripper-closure routine (incremental position commands with short delays) is executed to achieve reliable, compliant grasping that tolerates small placement inaccuracies (Fig. \ref{fig:grasping_pipeline}c).

\subsection{Mobile Navigation (SLAM \& Nav2)} 
\begin{figure}[h!]
    \centering
    % Left Image (a)
    \begin{minipage}{0.49\linewidth}
        \centering
        \includegraphics[height=5cm, width=\textwidth, keepaspectratio]{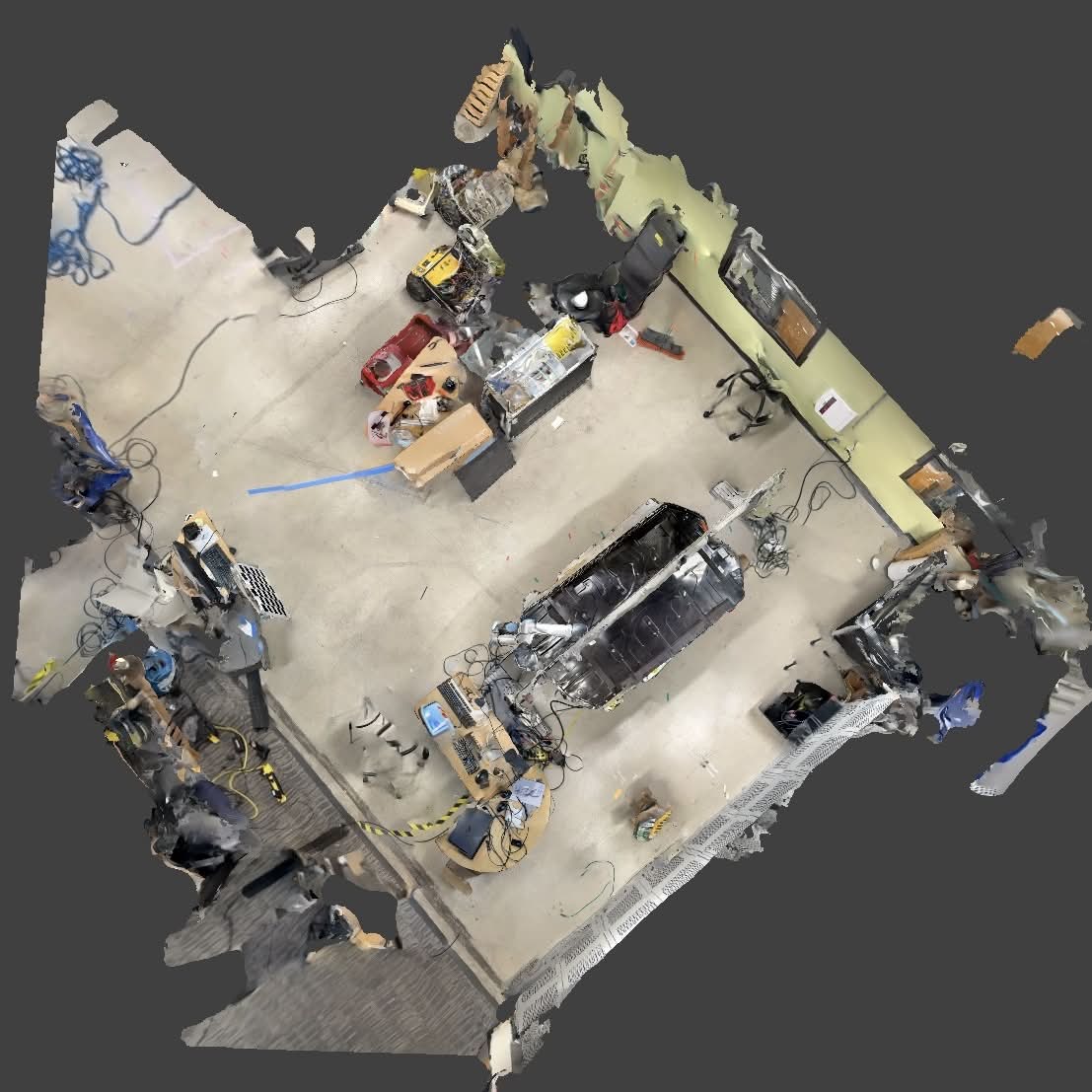}
        \centerline{(a) Ground truth}
        \label{fig:map_a}
    \end{minipage}
    \hfill
    % Right Image (b)
    \begin{minipage}{0.49\linewidth}
        \centering
        \includegraphics[height=5cm, width=\textwidth, keepaspectratio]{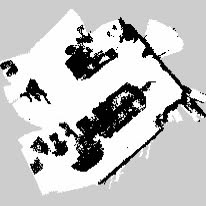}
        \centerline{(b) RTAB-Map}
        \label{fig:map_b}
    \end{minipage}
    
    \vspace{-5pt} % Minimal vertical space adjustment
    \caption{SLAM Navigation Representation. (a) ``Ground truth'' top down view of the laboratory (Scaniverse) and (b) corresponding 2D occupancy grid for autonomous navigation.}
    \label{fig:slam_results}
\end{figure}

For autonomous mobility, we implement a ROS~2 navigation pipeline that couples RGB-D perception with 2D costmap planning and control. Our current stack is designed around (1) a RealSense D435 sensor, (2) RTAB-Map \cite{labbe2019rtab} for visual odometry and SLAM, and (3) ROS~2 Nav2 for global planning and closed-loop execution. Fig. \ref{fig:slam_results} shows a 3D map of a test environment recorded using the \emph{Scaniverse} iPhone as a ground truth as well as a 2D occupancy map constructed using on-board sensors and RTAB-Map to the right.

We localize against a previously recorded map by loading an RTAB-Map database and running in localization-only mode (incremental mapping disabled). RTAB-Map consumes synchronized RGB, aligned depth, and camera intrinsics, and fuses visual odometry with wheel odometry (\texttt{/odom}) to produce a stable pose estimate in the \texttt{map} frame. This separation (map building vs.\ deployment-time localization) improves repeatability and reduces drift during long autonomous runs. 
 
Navigation is executed with standard Nav2 servers. A \texttt{map\_server} loads the static occupancy grid (Fig. \ref{fig:slam_results}b), the \texttt{planner\_server} generates global paths, and the \texttt{controller\_server} tracks paths to produce velocity commands on \texttt{/cmd\_vel}. For debugging and phased deployment, the controller can be toggled on/off, allowing map/localization validation before enabling closed-loop motion. 

Nav2 outputs standard \texttt{/cmd\_vel}. We forward these commands to the mobile base through a bridge process that supports linear/angular scaling and optional direction inversion. This isolates drivetrain-specific details from Nav2 and enables rapid adaptation across base variants without modifying planners or controllers. 

\subsection{VR Teleoperation Architecture}

To facilitate bimanual control, we implemented a Virtual Reality (VR) teleoperation system. Primarily, it provides an intuitive, human-in-the-loop interface that allows empirical validation of the kinematic workspace and dexterity of the bimanual mobile manipulator (Fig. \ref{fig:task_demos} and \ref{fig:vr_combined}). Secondarily, and crucially for modern embodied AI research and education, the system functions as a high-fidelity data collection pipeline. By recording synchronized streams of the operator's control inputs and the robot's sensor data (e.g. ego-centric camera feeds and joint states), the framework generates rich multimodal demonstration datasets required for large-scale imitation learning.

Building up on the Open-TeleVision architecture \cite{cheng2024tv}, our teleoperation framework processes both bare-hand tracking (wrist poses and pinch gestures) and standard VR controller actions. These inputs are dynamically mapped to the dual SO-101 manipulators using the real-time Inverse Kinematics (IK) solver that reliably operates at control frequencies up to 60 Hz, guaranteeing low-latency actuation. To provide an immersive telepresence experience, the robot's onboard camera feeds are projected directly into the operator's Head-Mounted Display (HMD). Furthermore, our framework is hardware-agnostic, supporting cross-platform deployment on headsets such as the Meta Quest, Apple Vision Pro, and Pico 4 Ultra. 

% EXPERIMENTAL EVALUATION
\section{Experimental Evaluation}
We evaluate: (1) structural properties and resulting manipulation capabilities, (2) computational benchmarking, (3) system stress testing, and (4) full-stack functional validation.

\subsection{Structural Rigidity and Load Capacity Analysis}
To evaluate our hybrid topology, we compared three printing strategies: Baseline (2-wall, 15\% infill), High-Infill (2-wall, 50\%), and High-Shell (4-wall, 15\%).

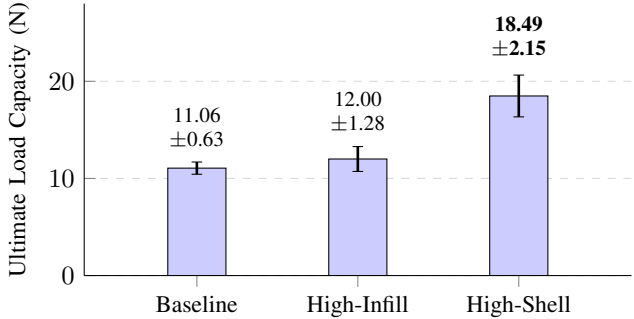
\begin{figure}[h]
\centering

% 1. The Comparison Table (Layered Top)
\renewcommand{\arraystretch}{1.2}
\resizebox{\columnwidth}{!}{
\begin{tabular}{lcccc}
\toprule
\textbf{Profile} & \textbf{Walls} & \textbf{Infill} & \textbf{Mass} & \textbf{Deflection ($\Delta z$)} \\
\midrule
Baseline & 2 & 15\% & 758g & 2 mm \\
High Infill & 2 & 50\% & 996g & 2 mm \\
\textbf{High Shell (Ours)} & \textbf{4} & \textbf{15\%} & \textbf{838g} & \textbf{2 mm} \\
\bottomrule
\end{tabular}
}

\vspace{4mm} % Precise vertical spacing

% 2. The Failure Force Histogram (Layered Bottom)
\begin{tikzpicture}
\begin{axis}[
    ybar,
    width=\columnwidth, 
    height=5.2cm, 
    bar width=22pt,
    enlarge x limits=0.35, 
    ylabel={Ultimate Load Capacity (N)},
    ylabel style={font=\small},
    symbolic x coords={Baseline, High-Infill, High-Shell},
    xtick=data,
    xticklabel style={font=\small},
    ymin=0, ymax=28, 
    axis lines*=left,
    ymajorgrids=true,
    grid style={dashed,gray!30}
]

\addplot[
    fill=blue!20, draw=black,
    error bars/.cd, y dir=both, y explicit, 
    error bar style={line width=1pt}
] 
coordinates {
    (Baseline, 11.06) +- (0, 0.63)
    (High-Infill, 12.00) +- (0, 1.28)
    (High-Shell, 18.49) +- (0, 2.15)
};

% Data Labels positioned for maximum clarity
\node[font=\footnotesize, align=center, anchor=south] at (axis cs:Baseline, 12.2) {11.06\\$\pm$0.63};
\node[font=\footnotesize, align=center, anchor=south] at (axis cs:High-Infill, 13.8) {12.00\\$\pm$1.28};
\node[font=\footnotesize, align=center, anchor=south] at (axis cs:High-Shell, 21.5) {\textbf{18.49}\\\textbf{$\pm$2.15}};

\end{axis}
\end{tikzpicture}

\caption{Structural Performance Characterization. Comparison of link topologies across rigidity (top) and ultimate load capacity (bottom, $n=5$).}
\label{fig:mechanical_combined}
\end{figure}

We first measured structural rigidity by applying a 100g static load to the cantilevered arm at 18.6cm extension and measuring the downward deflection. As shown in the table in Fig. \ref{fig:mechanical_combined}, the High-Shell configuration matched the deflection of the High-Infill profile (2.0mm) while reducing total mass by 15.8\% (838g), confirming that perimeter reinforcement is the superior strategy for maximizing stiffness-to-weight ratio.

Additionally, we characterized the maximum load capacity ($n=5$) of different topologies using a digital force gauge (SHIMPO FGV-10XY) to measure the force required to stall the actuator gearbox (rated at 1Nm, 3Nm peak). The arm was fixed in a cantilevered position while a downward vertical force was incrementally applied until reaching the actuator's holding torque limit, manifested by either audible motor stalling or internal gearbox failure (Fig. \ref{fig:robotic_arm_full}). As illustrated in the box-plots in Fig. \ref{fig:mechanical_combined}, High-Shell achieved a mean System Yield Force of $18.49 \pm 2.15$\,N, a 67.1\% increase over the Baseline ($11.06 \pm 0.63$\,N). These results validate that increasing wall density significantly improves functional load capacity ($\approx$1\,kg) for grasping tasks while optimizing material usage.

\subsection{Grasp Reliability Validation}
\begin{figure}[h!] 
    \centering
    
    % --- Row 1: Photos (a, b, c) - SLIGHTLY LARGER ---
    \begin{minipage}{0.325\columnwidth}
        \centering
        \includegraphics[width=\linewidth]{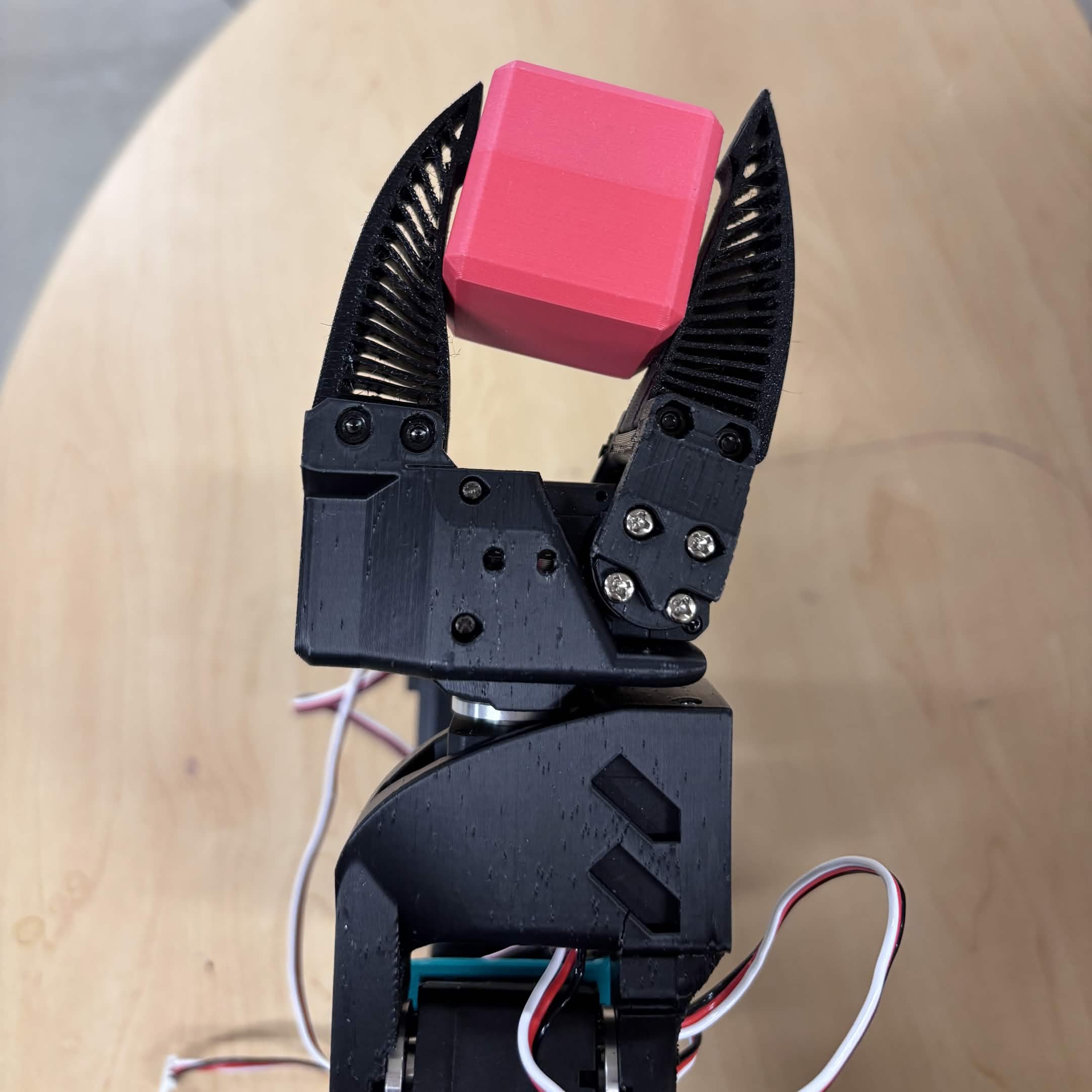}
        \small (a) Rigid
    \end{minipage}\hfill
    \begin{minipage}{0.325\columnwidth}
        \centering
        \includegraphics[width=\linewidth]{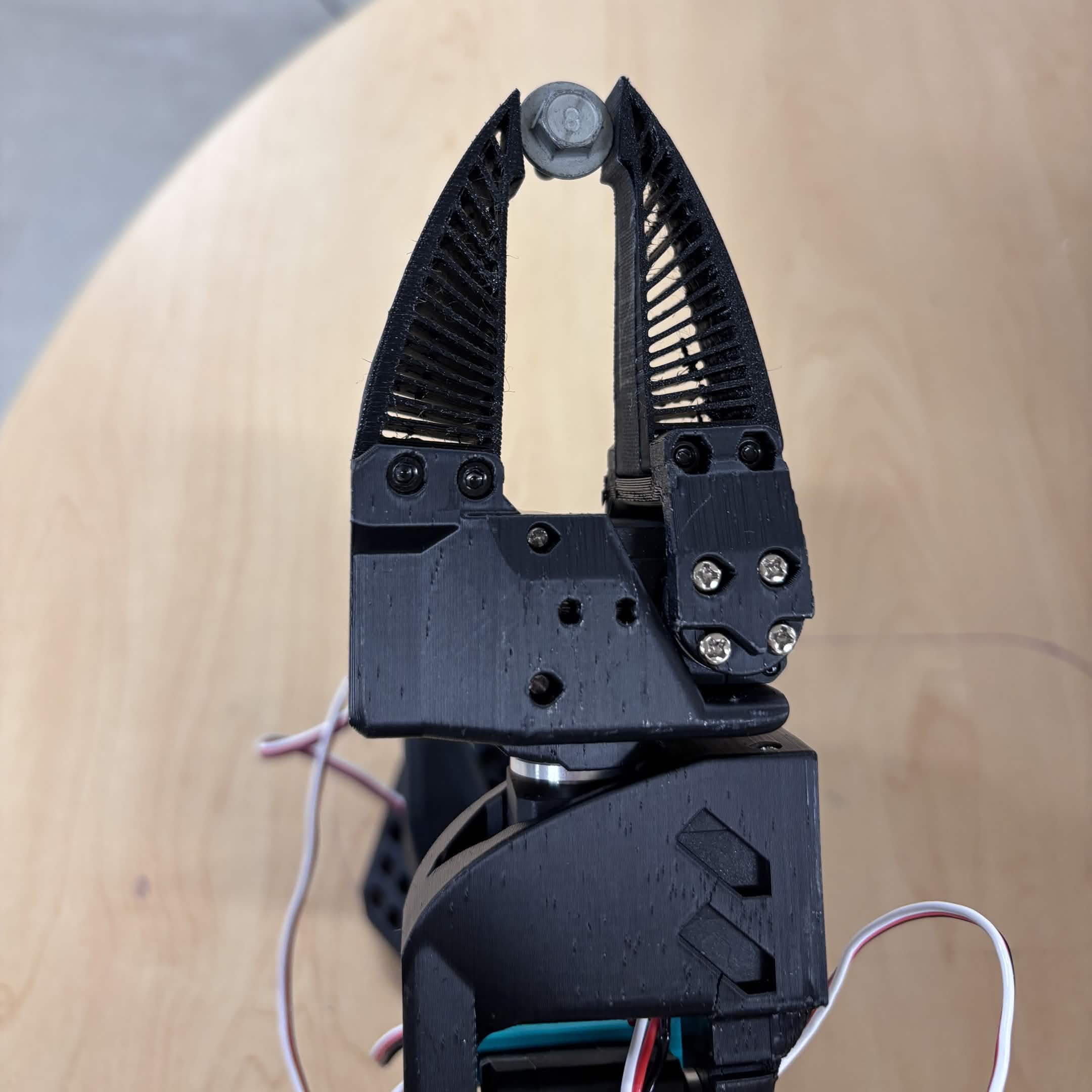}
        \small (b) Precision
    \end{minipage}\hfill
    \begin{minipage}{0.325\columnwidth}
        \centering
        \includegraphics[width=\linewidth]{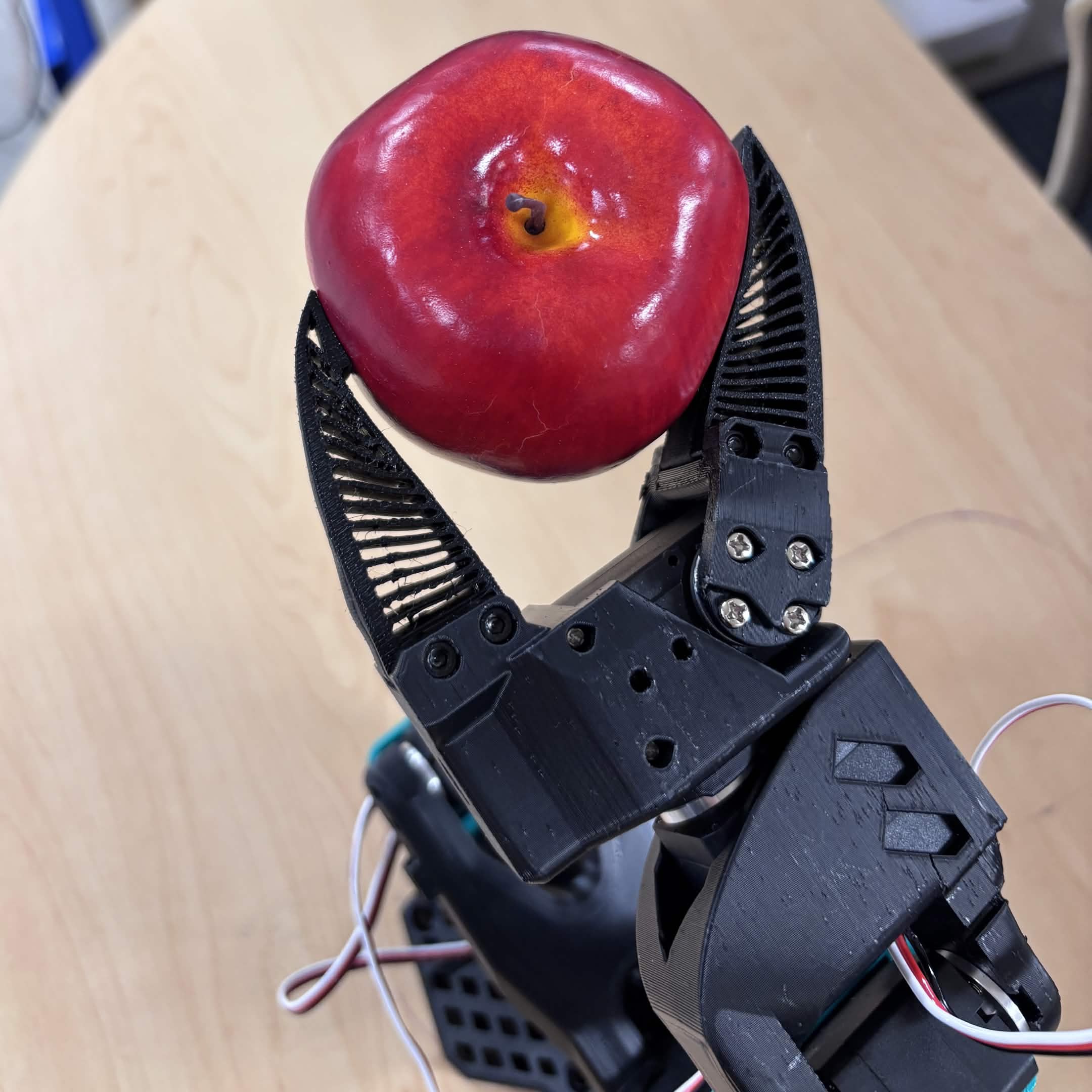}
        \small (c) Organic
    \end{minipage}

    \vspace{4mm} % Space between photos and matrix

    % --- Row 2: TikZ Matrix (d) ---
    \begin{minipage}{\columnwidth}
        \centering
        \resizebox{0.95\linewidth}{!}{%
        \begin{tikzpicture}[x=0.5cm, y=0.5cm]
            \tikzstyle{success}=[fill=green!60!black!20, draw=white, line width=0.8pt]
            \tikzstyle{fail}=[fill=red!70!black!20, draw=white, line width=0.8pt]
            
            \node[anchor=north] at (7.5, -1.25) {\textbf{Trial Index ($N=15$)}};
            
            \node[anchor=east] at (-0.2, 4.5) {Power Drill (858g)};
            \node[anchor=east] at (-0.2, 3.5) {Screwdriver (47g)};
            \node[anchor=east] at (-0.2, 2.5) {3D printed Cube (24g)};
            \node[anchor=east] at (-0.2, 1.5) {Plastic Apple (17g)};
            \node[anchor=east] at (-0.2, 0.5) {Mini water bottle (252g)};
            
            \foreach \x in {1,...,15} \node[anchor=north] at (\x-0.5, 0) {\small \x};
            
            \foreach \x in {0,...,14} \draw[success] (\x, 4) rectangle ++(1,1);
            \foreach \x in {0,...,14} \draw[success] (\x, 3) rectangle ++(1,1);
            \foreach \x in {0,...,14} \draw[success] (\x, 2) rectangle ++(1,1);
            \foreach \x in {0,1,2,3} \draw[success] (\x, 1) rectangle ++(1,1);
            \draw[fail] (4, 1) rectangle ++(1,1); 
            \node[text=red!80!black] at (4.5, 1.5) {\textbf{X}};
            \foreach \x in {5,...,14} \draw[success] (\x, 1) rectangle ++(1,1);
            \foreach \x in {0,...,14} \draw[success] (\x, 0) rectangle ++(1,1);
            
            \begin{scope}[shift={(2.5,-2.8)}]
                \draw[success] (0,0) rectangle (0.8, 0.4);
                \node[anchor=west] at (0.8, 0.2) {\small Success};
                \draw[fail] (3.5,0) rectangle (4.3, 0.4);
                \node[anchor=west] at (4.3, 0.2) {\small Failure (Slip)};
            \end{scope}
        \end{tikzpicture}}
        \centering
        \small (d) Grasp Consistency Matrix
    \end{minipage}

    \caption{Grasp Performance and Reliability. (a-c) Compliant mechanical adaptation. (d) Experimental results demonstrating a 98.7\% success rate.\vspace{-10pt}
    \label{fig:combined_grasp_results}}
\end{figure}
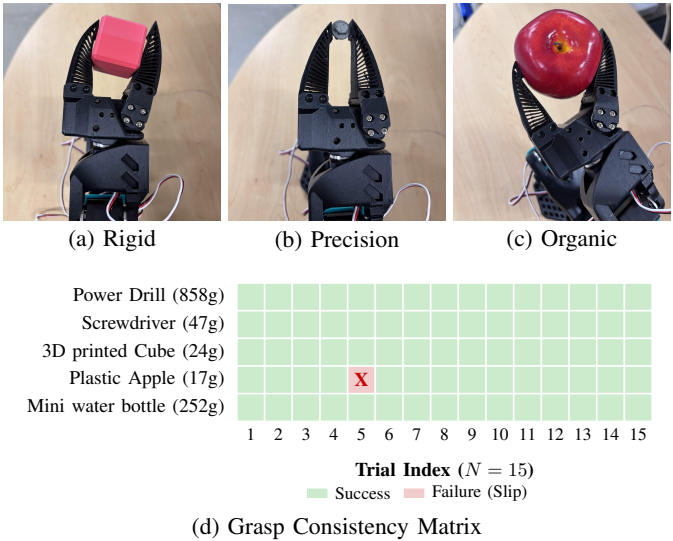

We evaluated gripper reliability through pick-and-place trials ($N=75$, 15 per object) across five objects (Fig. \ref{fig:combined_grasp_results}) spanning various sizes and weights (17--858g), utilizing the validated 4-wall design. The system achieved a 98.7\% success rate. Compliant Fin Ray mechanics provided 100\% success on rigid objects (cube, bottle) by leveraging passive morphological adaptation to maximize contact area. The 858g drill was successfully lifted but showed marginal stability due to high inertial loads and low-friction surfaces, leading to slip during rapid accelerations. A single failure occurred with the plastic apple due to an improper initial grasp.

\subsection{Computational Benchmarking}
To evaluate system performance on edge hardware, we benchmarked three state-of-the-art manipulation models within an end-to-end control loop (encompassing camera capture, image preprocessing, batch construction, and policy inference). The evaluated models include Action Chunking with Transformers (ACT) \cite{act_aloha}\footnote{\url{https://huggingface.co/lerobot/act_aloha_sim_transfer_cube_human}}, Diffusion Policy \cite{diffusion_policy}\footnote{\url{https://huggingface.co/lerobot/diffusion_pusht}}, and the 450M-parameter SmolVLA \cite{smolvla}\footnote{\url{https://huggingface.co/lerobot/smolvla_base}}. 

All evaluations were conducted directly on the Jetson Orin Nano, configured in the maximum performance MAXN SUPER Mode. To optimize latency, models were run in half-precision (FP16) utilizing a fused Scaled Dot-Product Attention (SDPA) kernel \cite{dao2022flashattention}. Models were configured for chunked action execution \cite{act_aloha}, with varying action horizons ($H$) and execution prefixes ($K$). For both SmolVLA and Diffusion Policy, we fixed the number of sampling steps to $T$=10.

Based on our empirical measurements, we report compute-limited upper bounds for the system's operational frequencies. Assuming policy inference represents the primary bottleneck and streaming previously planned actions incurs negligible overhead, the maximum theoretical replanning frequency is defined as:

$$ f_{\text{replan}}^{\max} = \frac{1}{\bar{t}_{\text{e2e}}} $$

where $\bar{t}_{\text{e2e}}$ is the measured mean end-to-end latency. 

A detailed breakdown of the latency and theoretical frequency limits for each architecture is presented in Table \ref{tab:inference_benchmarks}. Note that these calculations exclude downstream physical limitations, such as servo constraints and serial communication overhead, and thus represent theoretical upper bounds rather than guaranteed closed-loop performance.

\begin{table}[hbt!]
\centering
\caption{Inference Benchmarks on Jetson Orin Nano (MAXN SUPER Mode)}
\label{tab:inference_benchmarks}
\vspace{-10pt}
\begin{tabular}{l c c c c c}
\toprule
\textbf{Model} & \textbf{$H$} & \textbf{$K$} & \textbf{$T$} & $t_{\text{e2e}}$ (ms) & $f_{\text{replan}}^{\max}$ \\ \midrule
ACT & 100 & 50 & - & 36.0 $\pm$ 0.9 & 27.8 \\
Diffusion & 20 & 10 & 10 & 539.6 $\pm$ 0.1 & 1.8 \\
SmolVLA & 20 & 10 & 10 & 713.8 $\pm$ 9.3 & 1.4 \\ \bottomrule
\end{tabular}
\\[1ex]
{\raggedright \small \emph{Note:} E2E latency $t_{\text{e2e}}$ covers the loop from camera acquisition to action output. H=Horizon, K=Prefix, T=Sampling steps.\par}
\end{table}

Crucially, our architectural comparison reveals that the primary computational bottleneck on edge devices stems from the iterative action experts and the requisite denoising steps they demand. The data demonstrates that visual-semantic processing in SmolVLA introduces only minor overhead compared to a standard Diffusion Policy, proving that high-parameter semantic heads are not the only prohibitive factor for high-frequency robotic control.

\subsection{System Stress Tests} 
We subjected the system to high-duty cycle stress tests, simulating concurrent multi-axis movements (e.g., bimanual arm actuation and neck rotation) to evaluate power stability and thermal management, monitored via oscilloscope and onboard telemetry.

In the original shared-bus design, these loads induced a current surge and voltage collapse from 12.2V to 306mV at $t=19$s, necessitating a full power-cycle for recovery. This failure mode repeated across all 5 trials. The revised Tri-Bus architecture maintained stable voltage (mean 12.0V, variance $\leq 0.1V$) under identical conditions for 1-minute trials ($n=5$), eliminating transients and brownouts.

Thermal tests confirmed no Jetson throttling after 30 minutes of continuous operation at peak inference load while running the 450M-parameter SmolVLA model \cite{smolvla} (maximum GPU temperature $54.62^{\circ}\mathrm{C}$), validating the passive ducting design for untethered deployment. Furthermore, the robot maintained operational stability with a battery discharge of only 5\% over 30 minutes of full functional load. These results address common voltage instability in low-cost systems \cite{energy_sources}.

% \subsection{Full-Stack Functional Validation}
% We evaluated the integrated system through three applications: teleoperation, autonomous mapping, and mobile manipulation, demonstrating untethered performance.

% \subsubsection{Teleoperation Dexterity}

\subsection{Teleoperation Dexterity}

\begin{figure}[h!]
    \centering
    \begin{minipage}{0.5\linewidth}
        \centering
        \includegraphics[height=5cm, width=\textwidth, keepaspectratio]{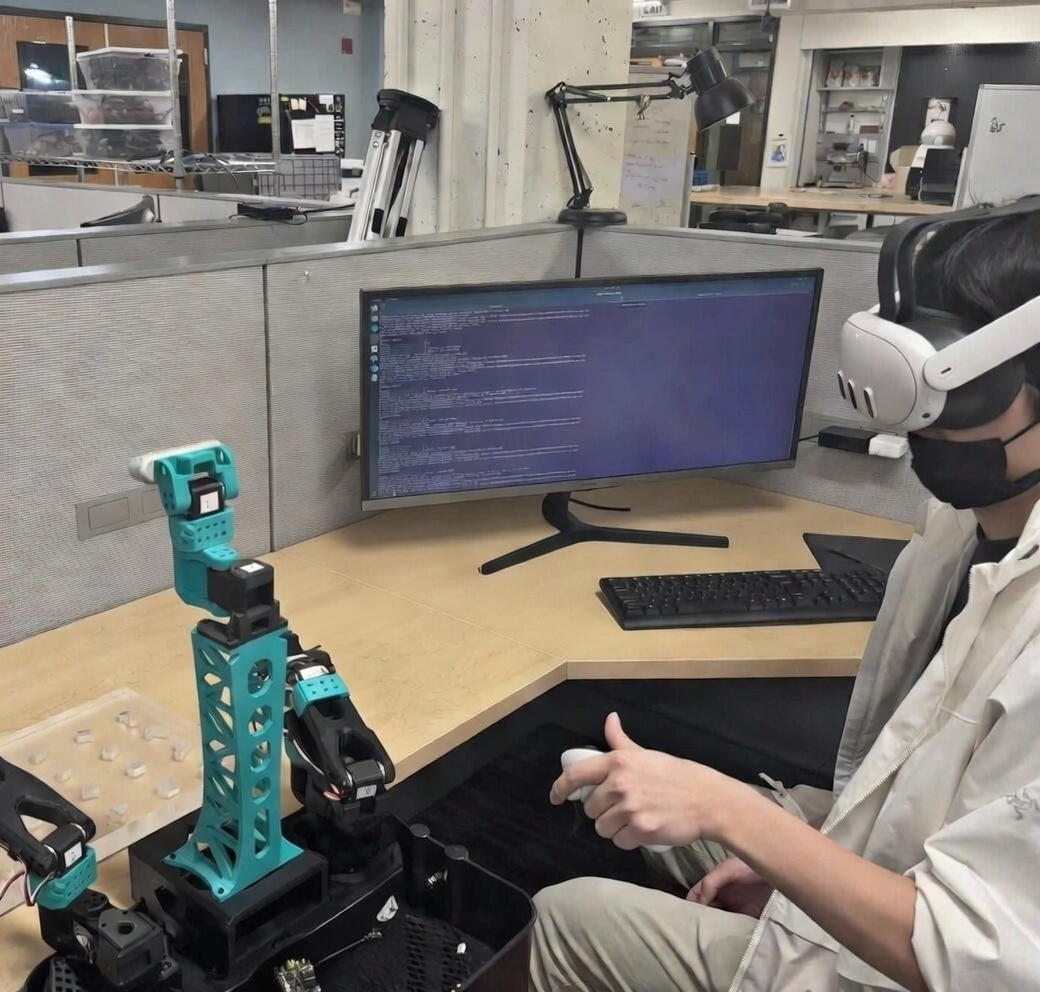}
    \end{minipage}
    \hfill
    \begin{minipage}{0.48\linewidth}
        \centering
        \includegraphics[height=5cm, width=\textwidth, keepaspectratio]{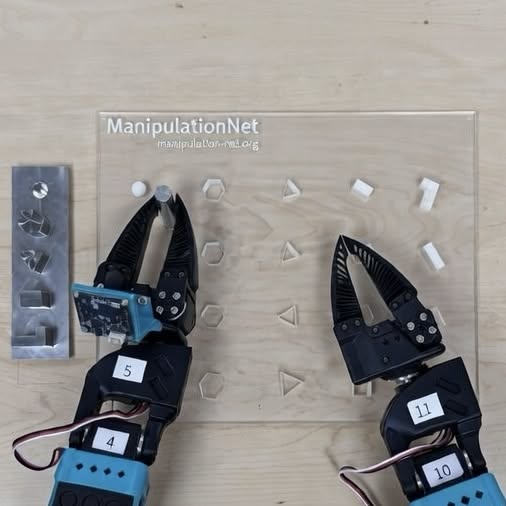}
    \end{minipage}
    
    \vspace{-3pt}
    \caption{VR Teleoperation Setup: (Left) The operator interacting with the system; (Right) a detailed view of the ManipulationNet peg-in-hole board.}
    \label{fig:vr_combined}
\end{figure}

We evaluated the effectiveness of the teleoperation interfaces for precise bimanual manipulation using a peg-in-hole insertion task, requiring the insertion of a 12mm peg into a 15mm hole (Fig. \ref{fig:vr_combined}). The study measured success rate (defined as full insertion without dropping the workpiece), task completion time, and the number of required correction attempts. To account for user variability, we conducted the study with three participants of varying robotics experience, totaling 25 trials across three modalities. Aggregated results for intuitiveness, ergonomics, and temporal performance are reported in Table \ref{tab:vr_benchmark}.

\begin{table}[h]
\centering
\caption{Teleoperation Performance on High-Precision Peg-in-Hole Task ($N=25$ total trials)}
\vspace{-10pt}
\label{tab:vr_benchmark}
\renewcommand{\arraystretch}{1.2}
\resizebox{\columnwidth}{!}{
\begin{tabular}{lccc}
\toprule
\textbf{Modality} & \textbf{Intuitiveness} & \textbf{Ergonomics} & \textbf{Time (s)} \\
\midrule
VR Controller\textsuperscript{*} & 2.67 & 2.33 & 48.22 \\
Hand Tracking\textsuperscript{*} & 1.33 & 1.67 & 68.30 \\
Xbox Controller Baseline & 1.50 & 3.67 & 248.73 \\
\bottomrule
\end{tabular}
}
\begin{flushleft}
\small \emph{Note:} \textsuperscript{*}Utilized the Meta Quest 3 AR system. Intuitiveness and Ergonomics represent mean scores on a 5-point Likert scale (1: Poor, 5: Excellent). Time represents the mean duration of trials.
\end{flushleft}
\end{table}

While both VR-based modalities reduced average task completion time by over 70\% compared to the joypad baseline (248.73s), the choice of input method proved critical for reliability. Using Meta Quest 3 (MQ3) with dedicated controllers achieved the highest performance with an 80.6\% time reduction (48.22s), whereas Hand Tracking---despite a 72.5\% reduction (68.30s)---suffered from significantly higher failure rates. These findings highlight a distinct trade-off: spatial hand mapping is more intuitive than coordinate-based joysticks, yet optical tracking latency and occlusion issues bottleneck high-precision bimanual tasks compared to physical controllers.

\section{Discussion and Limitations}
The robot's fully untethered operation, 1kg payload and 40cm reach have the potential to demonstrate simple home automation and manufacturing use cases (Fig. \ref{fig:battery_disassembly} and \ref{fig:task_demos}). Here, the platform can introduce a new baseline in terms of capability and cost.

Its low cost makes it suitable as an educational platform that can be shared among student groups. Finally, its modular architecture makes it possible to teach even larger classes in which students share a large number of arms, mobile bases, and a few complete xLeRobots, allowing to implement a curriculum such as described in \cite{curriculum}.

In addition to teaching basics ranging from vision and inverse kinematics to machine learning, students can get first-hand experience in fabricating their own hardware. Here, learning goals would not be limited to coding, but experiencing design trade offs in electrical and mechanical design such as described here, as well as experimenting with entirely novel designs including adding tactile sensors, additional cameras, or articulated hands, e.g.
Several limitations present opportunities for future work. The Jetson Orin Nano's limited CUDA kernels restricts high-frequency multi-modal transformers while a Jetson Thor (40-130W) exceeds the power budget, prompting exploration of tiered compute architectures or additional power supplies, including using an additional Anker power supply. Additionally, hobby-grade servos, while cost-effective, exhibit mechanical sensitivity; upgrading to industrial-grade servo motors would enhance durability and payload, but significantly increase cost by a factor 3-10x. Finally, the current configuration favors table-top tasks---adding a vertical linear axis like in the AhaRobot \cite{aharobot} will enable ground-level manipulation.

% --- CONCLUSION ---
\section{Conclusion}
This paper presents an untethered evolution of the XLeRobot platform that combines bimanual mobile manipulation with fully onboard GPU compute at a total system cost below \$1300. By jointly addressing mechanical stiffness, power integrity, and embedded compute integration, the proposed system transforms a previously tethered research platform into a self-contained mobile manipulation system capable of teleoperation, SLAM-based navigation, and vision-driven manipulation using onboard neural network inference. The tri-bus power architecture and improved structural design significantly increase system robustness while maintaining the accessibility and low cost that make the LeRobot ecosystem attractive for research and education.

Beyond the specific hardware design, this work highlights how modern embedded GPUs enable untethered robot learning systems at price points accessible to individual labs, classrooms, and distributed data-collection efforts. The resulting platform provides a practical testbed for emerging paradigms such as vision-language-action models, imitation learning, and large-scale teleoperation data collection on mobile manipulators. By lowering both the financial and technical barrier to entry, the platform can help standardize hardware for robot learning and facilitate reproducible research across institutions.

Future work will explore tighter integration with learning pipelines in the LeRobot ecosystem, including onboard policy deployment, large-scale teleoperation data collection, and benchmarking of learning-based mobile manipulation tasks. We hope that the open design and low cost of the platform will encourage community contributions and accelerate experimentation in embodied AI and robot learning.

%  TEMPORARILY REDACTED FOR DOUBLE BLIND. DO NOT REMOVE !!!

\section*{Acknowledgment}
The authors would like to thank Amine Boubrit for his contributions to the inverse kinematics (IK) algorithm development, and Lyle Antieau for his valuable technical support to our platform. We also extend our gratitude to the open-source communities of LeRobot and XLeRobot for providing the foundational frameworks for this research. This work has been supported by ARPA-E contract DE-AR0001966, "Robust Robotic Disassembly of EV Battery Packs using Open-World Vision Language Models and Symbolic Replanning.", we are grateful for this support. %Generative AI (Gemini AI) was utilized for structural refinement, grammatical proofreading, and the generation of visualization code (Tikz) throughout this manuscript.

\bibliographystyle{IEEEtran}
\bibliography{xlerobot}

\end{document}